\documentclass[conference,a4paper]{IEEEtran}
\ifCLASSINFOpdf
\else
\fi
\IEEEoverridecommandlockouts
\usepackage{amssymb}
\usepackage{graphicx}
\usepackage{multirow}
\usepackage{caption}
\usepackage{url}
\usepackage{pifont}
\usepackage{array,booktabs}

\usepackage{algorithm}
\usepackage[noend]{algpseudocode}

\usepackage[table]{xcolor}
\usepackage{hhline}

\definecolor{RobGreen}{RGB}{146,232,156}
\definecolor{RobYellow}{RGB}{232,232,146}
\definecolor{RobRed}{RGB}{232,146,146}

\usepackage[absolute]{textpos}
\usepackage{blindtext} % Just to have some dummy text

\begin{document}
\title{Hellinger Distance Trees for Imbalanced Streams}

\author{
    \IEEEauthorblockN{R. J. Lyon, J. M. Brooke, J. D. Knowles}
    \IEEEauthorblockA{
    School of Computer Science\\
	University of Manchester\\
	Manchester, UK\\
    }
    \and
    \IEEEauthorblockN{B. W. Stappers}
    \IEEEauthorblockA{
    Jodrell Bank Centre for Astrophysics\\
	University of Manchester\\
	Manchester, UK\\
    }
    
    \IEEEcompsocitemizethanks{
    \IEEEcompsocthanksitem This work was supported by grant EP/I028099/1 for the University of Manchester Centre for Doctoral Training in Computer Science, from the UK Engineering and Physical Sciences Research Council (EPSRC).}% <-this % stops a space
%\thanks{Manuscript created April 1, 2013; revised April 3, 2013.}
}

\maketitle

\begin{abstract}
Classifiers trained on data sets possessing an imbalanced class distribution are known to exhibit poor generalisation performance. This is known as the imbalanced learning problem. The problem becomes particularly acute when we consider incremental classifiers operating on imbalanced data streams, especially when the learning objective is rare class identification. As accuracy may provide a misleading impression of performance on imbalanced data, existing stream classifiers based on accuracy can suffer poor minority class performance on imbalanced streams, with the result being low minority class recall rates. In this paper we address this deficiency by proposing the use of the Hellinger distance measure, as a very fast decision tree split criterion. We demonstrate that by using Hellinger a statistically significant improvement in recall rates on imbalanced data streams can be achieved, with an acceptable increase in the false positive rate.
\end{abstract}



% \fi
\IEEEpeerreviewmaketitle

\section{Introduction}
\label{sec:Introduction}
Imbalanced data sets are characterised by a skewed class distribution, which typically favours a single majority class, over one or more minority classes. Nearly all datasets are imbalanced to some degree, however the extent to which they are imbalanced varies greatly. Those data sets which exhibit the largest imbalances present significant problems for inductive learners, since algorithms trained on such data often exhibit poor generalisation capabilities on test data. The cause of such poor performance has been the subject of extensive study by the research community since the early 2000's \cite{AAAI_ws:2000:jp,ICML_ws:2003:nc,Chawla:2004:jnka}. The problem is now fairly well understood, and a number of useful approaches have been developed that generally seem to improve classifier performance on imbalanced datasets (see related work in Section \ref{sec:Related}). However these approaches are intended for use on static datasets. These are data sets of a known fixed size $n$. In situations where the size of the dataset is indeterminable as with data streams, these approaches become unsuitable, either due to their reliance on knowing $n$ a priori, or their higher computational requirements. In this paper we combine the Hellinger distance used previously in some static imbalanced learners \cite{Cieslak:2008:cn} with a stream classifier based on the Hoeffding bound. Together these can be used to produce a skew insensitive decision tree split criterion that substantially increases minority class recall rates. Our goal is to use this approach to extract rare class instances from a heavily imbalanced data stream arising from the pulsar astronomy domain.
\begin{textblock}{170}(30,280)
\noindent Preprint submitted to 22nd International Conference on Pattern Recognition. Received December 19, 2013.
\end{textblock}

\section{Motivation}
\label{sec:Motivation}
The motivation for this work is provided by the development of the world's largest radio telescope, the Square Kilometre Array (SKA). The SKA will comprise thousands of individual 15 metre receiving dishes, which together form an instrument capable of producing data at a rate of many TB/s \cite{Smits:2009dc}. When used to survey the sky for periodic signals produced by a rare type of stellar object, the radio pulsar, such a high data capture rate presents a number of significant computational challenges and barriers to discovery. Our principal dilemma centres on determining how to separate the small number of scientifically interesting signal detections made by the SKA, from the large volume of negative detections arising from radio frequency interference (RFI) and background noise. The data to be processed is an intrinsically imbalanced collection of overwhelmingly negative detections requiring separation, which maps neatly to a traditional machine learning classification problem. However since the SKA's high data capture rate makes it infeasible for all data to be stored permanently (due to financial restrictions), the classification of received data will likely have to be performed on arrival, in close to real-time. The continual arrival of new signals for processing is analogous to a stream of data flowing from the telescope, through a computational processing pipeline. Thus we view the identification of rare signals in SKA data as a stream classification problem.\IEEEpubidadjcol
 
%Such streams may seem somewhat unusual, however in practice large imbalances do occur in many domains. A prime example is provided by the Atlas experiment \cite{ATLAS:2008:bc} conducted at the Large Hadron Collider (LHC). The data produced by this experiment can be viewed as a high throughput stream of signal and non-signal events. In this case too, only a small fraction of the events are of interest, meaning these must be filtered in order to isolate those that may lead to a discovery. Other general areas which are characterised by imbalanced data sets include medical diagnosis and biological data analysis \cite{Guyon:2002:IS,Mazurowski:2008:pa,Radivojac:2004:nc}, analysis of imagery \cite{Kubat:1998:ML,Casasent:2004:xw}, fraud detection \cite{Chan:1998:ss,Phua:2004:ca,Wei:2013:wl}, and text classification \cite{Mladenic:1999:FS,Forman:2003:EE,Zheng:2004:FS}.\IEEEpubidadjcol

\section{Problem Definition}
\label{sec:ProblemDefinition}
We are concerned with the identification of rare class instances within a data stream emerging from a radio telescope's data processing pipeline, which possesses a class distribution of approximately +1:-10,000\footnote{Estimation based on data obtained during a past pulsar survey \cite{Lorimer:2006ha}.}. Each instance in the stream is a pulsar candidate, a summary description of each signal exhibiting specific characteristics of interest. Our aim is to build a binary classifier capable of separating candidates arising from noise or RFI (the negative majority class), from those generated by radio pulsars (the positive minority class). We define the input stream as $S={ \lbrace ...,(x_{i}),... \rbrace}$, $i=1,...,\infty$ which describes the candidates emerging from a pulsar search pipeline under a discrete time model. Each candidate in the stream $x_{i} \in X$ is defined as $x_i = \lbrace (x_{i}^{1}),...,(x_{i}^{m}) \rbrace$, where each $x_{i}^{j} \in \mathbb{R} $ for $j=1,...,m$ is a single summary statistic that describes some aspect of the signal represented by the candidate $x_{i}$. We specify binary candidate labels $C = \lbrace -1,1 \rbrace $, where $c_{i}$ is an individual label such that $c_{i} \in C$ (i.e. uninteresting =~-1, interesting~=~1). Our goal is to learn a function $f: X \mapsto C $ which maps each candidate to its correct label producing the set of labelled candidates. From these the set of positively labelled candidates can be obtained, which should be stored for expert inspection. Building a classifier to achieve this on a static data set is straightforward. We need only form a suitable training set and apply the classification model learned to a test set. In a streamed classification scenario there is no distinction between training and test sets, since all data resides in the stream. There are instead two general processing models used for learning.
\label{batch}
\begin{itemize}
\item {\textbf{Batch processing model:}} at time step $i$ a batch $b$ of $n$ unlabelled instances arrives, and is classified using some model trained on batches $b_{1}$ to $b_{i-1}$. At time $i+1$ labels arrive for batch $b_{i}$, along with a new batch of unlabelled instances $b_{i+1}$ to be classified.
\item {\textbf{Incremental processing model:}} a single data instance arrives at time step $i$ defined as $x_{i}$, and is classified using some model trained on instances $x_{1}$ to $x_{i-1}$. At time $i+1$ a label arrives for $x_{i}$, along with a new unlabelled instance $x_{i+1}$ to be classified.
\end{itemize}
We adopt the incremental processing model throughout this work since it neatly describes the arrival of each candidate. However the model must be adapted since the stream $S$ as described previously is completely unlabelled, given it is impossible to know a priori the correct label for each candidate\footnote{Accurate labels can only be obtained via re-observation, which is financially costly and time consuming.}. To accommodate this we prepend labelled training examples\footnote{There are some 2,000 pulsars known to science for which training candidates can be generated.} to the start of the stream, ensuring a classification model is built from instances $x_{1}$ to $x_{i-1}$. We refer to this process here as `off-line pre-training'. Furthermore, true class labels cannot be obtained for every instance at $i+1$ due to the candidate volume and the cost of analysis. In reality only instances receiving positive predictions will be labelled following expert analysis. This is simulated in the experiments described in Section \ref{sec:Experiments}, by assigning a varying proportion of instances in the stream their correct label at $i+1$. Thus we now consider the stream $S={ \lbrace ...,(x_{i},c_{i}),... \rbrace}$ where $c_{i}$ is a class label. For labelled instances in the stream originating from simulated expert analysis, $c_{i}$ will assume the value of a class label $\in C$, whilst for those without labels $c_{i}=0$ which our learning algorithms will ignore during training.

\section{Imbalanced Learning on Data Streams}
\label{sec:ImbalancedLearning}
\begin{figure}[htb]
	\centering
		\includegraphics[keepaspectratio,scale=0.125]{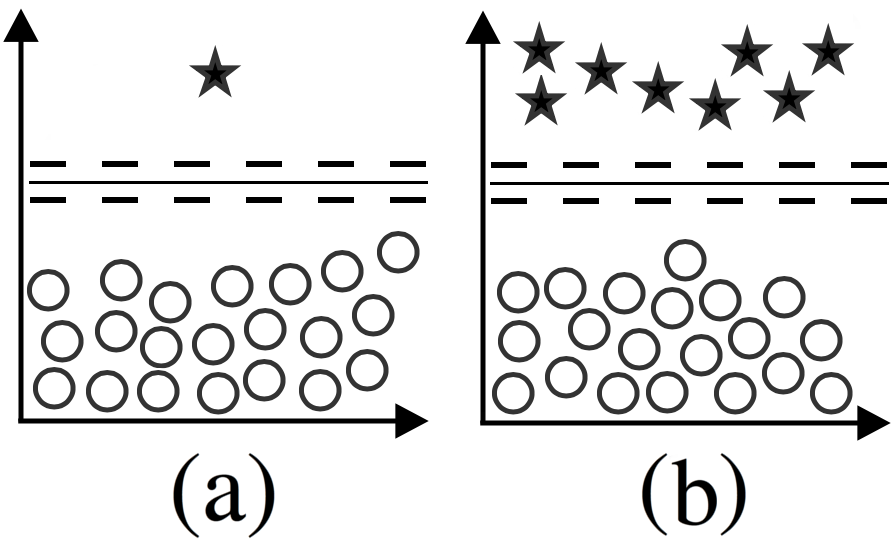}
		\caption[]{Training distribution (a) and test distribution (b). Here a skewed class distribution has no effect on classifier performance since a decision boundary can be induced which allows all instances to be classified correctly. Based on diagrams used in \cite{Haibo:2009:fb}.}
		\label{fig:SimpleBoundary}
		\vspace{-1.5em}
\end{figure}

\begin{figure}[htb]
	\centering
		\includegraphics[keepaspectratio,scale=0.125]{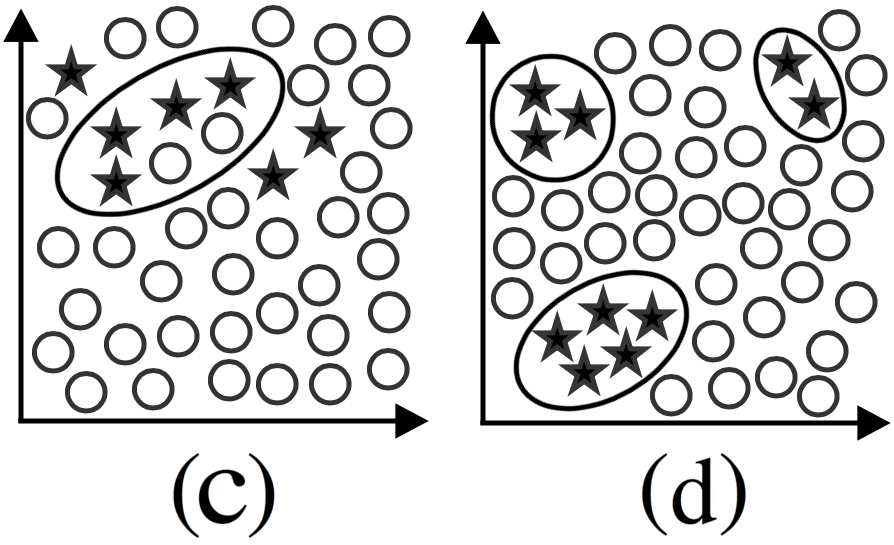}
		\caption[]{Class inseparability in (c) and small disjuncts in (d). Based on diagrams used in \cite{Haibo:2009:fb}.}
		\label{fig:TypesOfImbalance}
		\vspace{-1.5em}
\end{figure}

Training a classifier on an imbalanced dataset does not necessarily mean poor generalisation performance \cite{Galar:2012:fa}. If the training data are discriminative enough to separate the different classes in data space, then the classifier will perform well regardless of the imbalance. A contrived example of this is shown in Fig.\ref{fig:SimpleBoundary}. It demonstrates that the underlying data distribution alone is not the root cause of poor classification performance on imbalanced data \cite{Galar:2012:fa}. Rather it is three characteristics often possessed by imbalanced data sets that make it difficult for a classifier to separate the minority and majority classes. These are \textit{small sample size} \cite{Haibo:2009:fb,Galar:2012:fa}, \textit{class inseparability} \cite{Japkowicz:2002:ss,Galar:2012:fa} Fig.\ref{fig:TypesOfImbalance}(c), and \textit{small disjuncts} in Fig.\ref{fig:TypesOfImbalance}(d). Ultimately these characteristics conspire to make it difficult for a classifier to construct an optimal decision boundary leading to sub-optimal classifier performance.
\begin{textblock}{170}(30,280)
\noindent Preprint submitted to 22nd International Conference on Pattern Recognition. Received December 19, 2013.
\end{textblock}
Learning from non-static imbalanced data, is known to be a difficult and important problem yet to be overcome in data-mining research \cite{Xindong:2006:yq}. Whilst classifiers built upon imbalanced streams have strong accuracy overall, they possess poor performance on the minority class \cite{Wasikowski:2010:xc,Lyon:2013:jk}. Consider a two-class stream with an arbitrarily large imbalance i.e. $\lim_{+ \to 1}$ and $\lim_{- \to \infty}$. At any moment such a stream will possess only a small number of minority class instances. Thus a learner retrained over a random batch will do so on an intrinsically small sample of minority class instances, whilst an incremental learner will rarely be presented with a minority class example to learn from. This is the streamed version of the small sample problem. As the inductive bias of many machine learning algorithms is geared towards forming the simplest hypothesis possible, its consequences are severe: algorithms will typically predict the majority class for all instances \cite{Akbani:2004:sk}, particularly as it is easy to dismiss any positives seen as noise. In real world domains the small sample problem is worsened as streams are often either unlabelled or partially labelled. Any partial labelling of the stream will likely include only the majority class instances, and this becomes increasingly likely as the proportion of labelled instances decreases. Those minority class instances which would provide a more representative sample, are rarely, if ever labelled. Furthermore, real world classification tasks are non-trivial, i.e. there is some degree of class overlap. Even if these overlaps are caused by noise and are few in number, say $\frac{1}{1000}$ instances, if the minority class occurs infrequently enough in the stream i.e. $\frac{1}{10000}$, then in time this noise would drown out the minority concept leading to class inseparability. Indeed the noisy negative examples may quickly form new clusters in data space for which a classifier would need to develop small disjuncts to cover. In order to alleviate these problems we either need to provide a larger and more representative sample of minority class examples for learning, or alternatively, a new approach unaffected by\newpage\noindent class skew. In this work we have adopted the latter approach using the Hellinger distance measure as a skew insensitive split criterion for streamed decision trees. Note that we have not considered the effects of distributional drift (concept drift) here; we reserve such discussions for future work.
\section{Hellinger Distance}
\label{sec:Hellinger}
The Hellinger distance is a symmetric and non-negative measure of distributional divergence, related to the Bhattacharyya coefficient (BC)\footnote{Though Hellinger obeys the triangle inequality whereas the BC does not.} and the Kullback-Leibler (KL) divergence. Cieslak and Chawla proposed the use of the Hellinger distance as a decision tree split criterion for imbalanced data in \cite{Cieslak:2008:cn,Cieslak:2012:hr}. They were able to utilise the distance in this way by considering two distributions $P$ and $N$, to be the normalised frequencies of feature values in a binary classification scenario. In which case when $P=N$ there is no difference between the distributions therefore the distance is zero, whilst if the two are completely disjoint then the distance will be one. Cieslak and Chawla then use the notion of `affinity' between $P$ and $N$ as a decision tree split criterion. The aim is to split tree nodes on those features with minimal affinity i.e. maximal Hellinger distance. This approach is appealing since it enables the splitting of features based on how well they discriminate between the examples seen so far in the stream, rather than on the feature which describes the largest possible number of instances seen so far, as with information gain. Intuitively Hellinger is skew insensitive, since an abundance of examples of one class will only serve to bring its sample distribution closer to the real distribution. In which case if a feature is a good class discriminator, then irrespective of the balance it will remain as such. Theoretical evidence backing up this intuitive interpretation is supplied in \cite{Cieslak:2008:cn,Cieslak:2012:hr}. The Hellinger distance with respect to a single continuous feature for the two class case given in \cite{Cieslak:2008:cn}, where the feature is discretized into $b$ bins is defined as follows, 
\begin{eqnarray}\label{eq:Hellinger}
d_{H} (X_{+},X_{-}) &=&\sqrt{\sum_{j=1}^{b} \left(\sqrt{\frac{|X_{+j}|}{|X_{+}|}} -\sqrt{\frac{|X_{-j}|}{|X_{-}|}}\right)^{2}}\textrm{,}
\end{eqnarray}
where $X_{+}$ is the positive class count, $X_{-}$ the negative class count, and $X_{+j}$ the count of positives in bin $j$ (defined similarly for $X_{-j}$). The problem with this formulation is that it requires knowledge of the normalised frequencies of values for each feature. This requires $O(lfbc)$ memory for $l$ leaves in the tree, $f$ features, $b$ bins and $c$ classes. For some stream environments this could be too costly. However the Hellinger distance between two normal distributions $P$ and $N$ can be computed more straightforwardly, provided the mean and standard deviation of the feature is known. Given $P$ with mean $\mu_{1}$, variance $\sigma_{1}^{2}$ and standard deviation $\sigma_{1}$ with $N$ defined similarly; we can calculate the Hellinger distance as follows,
\begin{eqnarray}\label{eq:GaussianHellinger}
d_{H} (P,N) &=& \sqrt{1 - \sqrt{ \frac{2 \sigma_{1} \sigma_{2}} {\sigma_{1}^{2} + \sigma_{2}^{2}} } e^{ -\frac{1}{4} \frac{ (\mu_{1}-\mu_{2})^{2} } {\sigma_{1}^{2} + \sigma_{2}^{2}} }} \textrm{.}
\end{eqnarray}
The advantage of Eq. \ref{eq:GaussianHellinger} is that it requires we keep only simple statistics which describe the distribution of each feature requiring $O(lf\cdot 2c)$ memory. Thus we propose a new algorithm-level approach (as opposed to other approaches described in Section \ref{sec:Related}) which uses the Hellinger distance defined between two normal distributions, as a splitting criterion in Hoeffding trees (the Very Fast Decision Tree developed in \cite{Hulten:2001:MTD}). We call this the Gaussian Hellinger VFDT. Since this approach relies upon measuring distances between normal distributions, it will require a number of examples to be seen before a reasonable normal model can be computed for each feature. Consequently it does not initially possess the any-time property\footnote{Although this can be mitigated by a suitable period of `pre-training'.}.  Furthermore we are also assuming that features will be normally distributed. Note that in our implementation of  Eq. \ref{eq:GaussianHellinger} continuous attributes were discretized into 10 bins. Pseudo code for our implementation of the GH-VFDT is presented in Algorithm~1, based on the VFDT algorithm given in \cite{Hulten:2001:MTD}.
 \begin{algorithm}
\small
\caption{\footnotesize{Gaussian Hellinger Very Fast Decision Tree (GH-VFDT)}}
\begin{algorithmic}[1]
\Require An input dataset $	S=\lbrace ..., (x_{i},c_{i}),...\rbrace$, such that each $x_{i}$ is an instance and $c_{i}$ is its true class label. Also requires $\delta$ the confidence desired, and $\tau$ a parameter which prevents ties.
\Procedure {GH-VFDT}{$S,\delta,\tau$}
\State Let $DT$ be a decision tree with leaf $l_{1}$
\For{$i \leftarrow 1$ to $|S|$ } \Comment For each stream instance.	
	\State $l \leftarrow sort(x_{i},c_{i})$\Comment Sort instance $x_{i}$ to leaf $l$.	
	\State $k\leftarrow c_{i}$ \Comment Get class.
	\For{$j \leftarrow 1$ to $|x_{i}|$ } \Comment For each feature.
		\State update $\mu_{ijk}(l)$
		\State update $\sigma_{ijk}(l)$
	\EndFor
	\State Label $l$ with majority class of instances seen at $l$
	\If{ all examples seen at $l$ don't belong to same class }
	\State $F_{a} \leftarrow null$\Comment Best feature.
	\State $F_{b} \leftarrow null$\Comment 2nd best feature.
		\For{$j \leftarrow 1$ to $|x_{i}|$ } \Comment For each feature.
		\State $dist\leftarrow d_{H} (x_{i})$  \Comment From equation \ref{eq:GaussianHellinger}.
		\State $F_{a},F_{b} \leftarrow getBest(dist,x_{i})$
		\EndFor
		\State $\epsilon = \sqrt{\frac{R^{2} \textrm{ }ln(1 / \delta)}{2n}}$ \Comment Hoeffding bound.
		\If{ $d_{H} (F_{a})-d_{H} (F_{b}) > \epsilon$ or $\epsilon < \tau$ }
			\State Replace $l$ with new leaf that splits on $F_{a}$
			\For{each branch of split }
				\State Add new leaf $l_{m}$
				\For{$k \leftarrow 1$ to $|C|$ } \Comment For each class.
					\For{$j \leftarrow 1$ to $|x_{i}|$ } \Comment For each feature.
						\State $\mu_{ijk}(l_{m}) \leftarrow 0$
						\State $\sigma_{ijk}(l_{m})  \leftarrow 0$
					\EndFor
				\EndFor
			\EndFor
		\EndIf
	\EndIf
\EndFor
\State \Return $DT$
\EndProcedure
\end{algorithmic}
%\vspace{-.05em}
\end{algorithm}
\begin{textblock}{170}(30,280)
\noindent Preprint submitted to 22nd International Conference on Pattern Recognition. Received December 19, 2013.
\end{textblock}
\section{Experiments}
\label{sec:Experiments}
\subsection{Static Tests}
\label{sec:Experiments_A}
Static experiments attempted to reproduce the results for the static domain as described in \cite{Cieslak:2008:cn}. We applied $5\times2$-fold cross validation (CV)\footnote{Data split into two chunks, the first used for training the second for testing. This is then reversed (train on second, test on first) and repeated five times. 10-fold CV not used as it can give an elevated type 1 error rate \cite{Cieslak:2008:cn}.} in conjunction with the C4.4\footnote{An unpruned decision tree using Laplace smoothing at the leaves.}, C4.5, and the HDTree~\cite{Cieslak:2008:cn} algorithm which uses Eq.\ref{eq:Hellinger}; to four data sets used in \cite{Cieslak:2008:cn} (indicated in Table \ref{tab:data} with an asterisk). To evaluate our results we used metrics that give a representative impression of classifier performance on the minority class (e.g. the $\textrm{G-Mean} = \sqrt{\frac{TP}{TP+FN}\times \frac{TN}{TN+FP}}$). The area under the ROC curve (AUC), F-score and recall rate were also used. This test procedure was then repeated on larger datasets not used in \cite{Cieslak:2008:cn}. The purpose of running these experiments was to determine a baseline level of performance for static classifiers, against which stream performance could be compared.
\subsection{Stream Tests}
Data stream experiments were performed on a VFDT, the GH-VFDT presented in Algorithm~1, and a streamed implementation of the HDTree called HD-VFDT also written as part of this work. The HD-VFDT is the same algorithm as GH-VFDT, except Eq.\ref{eq:GaussianHellinger} has been replaced by Eq.\ref{eq:Hellinger}.  In this streamed scenario, the three largest data sets in Table \ref{tab:data} were shuffled (except for the temporally ordered pulsar data) and then randomly sampled in order to generate: 1) training sets containing 200 positive and 1000 negative instances to be used for pre-training, and 2) disjoint test sets possessing different levels of class imbalance and a varied proportion of labelled instances, simulating different levels of expert feedback. Each sampling permutation was treated as a single stream of data. For each algorithm, tests were repeated ten times for a given balance and labelling, allowing results to be averaged. An incremental test-then-train approach was adopted throughout the stream experiments, i.e. test on an instance, then train if the label is available. Results were evaluated predominately using the G-Mean. The popular AUC metric was not computed given that it is computationally expensive to do so for millions of instances\footnote{Whilst methods have been developed to provide AUC approximations for large data sets \cite{Bouckaert:2006:rr}, we do not make use of these here. Particularly as work in \cite{Hand:2009:dj} suggests the AUC is an unsuitable metric for use upon imbalanced data.}. To rigorously validate our stream results, one-factor analysis of variance (ANOVA) tests were performed, where the algorithm used was the factor. Tukey's Honestly Significant Difference (HSD) test was then applied to determine which results where statistically significant using $\alpha=0.01$.
\subsection{Data}
\label{sec:Data}
\begin{table}
\center
\tabcolsep=0.11cm
\scalebox{1.0}{
    \begin{tabular}{|l|l|l|l|c|c|}
    \hline
     \multirow{2}{*}{Dataset} & \multirow{2}{*}{Instances} & \multirow{2}{*}{Attributes / Type}  & \multirow{2}{*}{$\sim$Balance} & \multicolumn{2}{c|}{Distribution}\\\cline{5-6}
     & & & & + & -\\ \hline\hline
    Pulsar & 11,219,848 & 22 / Continuous & +1 : -7000 & 1,611 & 11,218,237\\
    Skin & 245,057 & 3 ~/ Discrete& +1 : -4 & 50,859 & 194,198\\
    MiniBoone & 130,065 & 50 / Continuous& +1 : -3 & 36,499 & 93,565\\
    Letter* & 20,000 & 16 / Continuous& +1 : -4 & 3,878 & 16,122\\
    Magic & 19,020 & 10 / Continuous& +1 : -2 & 6,688 & 12,332\\
    Pen* & 10,992 & 16 / Continuous& +1 : -9 & 1,055 & 9,937\\
    Statlog Landsat* & 6,435 & 36 / Continuous& +1 : -9 & 626& 5,809\\
    Statlog Image* & 2,310 & 19 / Continuous& +1 : -6 & 330 & 1,980\\\hline
    \end{tabular}
    }
\caption[]{Characteristics of the data sets used. Asterisks indicate data sets used in \cite{Cieslak:2008:cn}.}
\label{tab:data}
\vspace{-2.1em}
\end{table}
% Candidates are represented by 22 continuous numerical attribute which describes it in some way. The data set contains only 1,611 positive and 2,593 negative examples correctly labelled by human annotators. The remainder of the dataset is na\"{\i}vely assumed to be negative, though the number of actual positives incorrectly labelled in this dataset has been been estimated to be between 879-916 \cite{Lyon:2013:jk}. 
In total eight datasets were used for this work (see Table~\ref{tab:data}). The largest dataset consisted of pulsar candidates obtained during the HTRU survey \cite{Keith:2010bl}\footnote{This data is currently not publicly accessible.}. The remaining datasets were all obtained from the UCI machine learning repository. These include the Skin Segmentation, MiniBoone and the Magic Gamma Telescope datasets, describing binary classification problems. The Letter, Pen, Statlog Landsat and Statlog Image datasets on the other hand, describe multi-class problems. These were converted to binary datasets by labelling the smallest class within them as the minority class, and labelling all other instances as belonging to the majority class. For the Letter data set this meant grouping all vowel instances (without `y') to form the minority class, whilst for the Pen data set those instances with the class label `5' formed the minority class.
\subsection{Results}
\label{sec:Results}
On static data sets we find that the HDTree algorithm developed in \cite{Cieslak:2008:cn} achieves the best AUC values on 5 of the datasets used, as can be seen in Table \ref{tab:static}. On those that remain the HDTree has a joint best AUC value on two datasets, and is outperformed only once on the Magic dataset by the C4.4 algorithm. The AUC results obtained for the Letter, Pen, Landsat and Image datasets are similar to those reported in \cite{Cieslak:2008:cn}. Thus our results generally support those reported in \cite{Cieslak:2008:cn} for the HDTree. Though here we can see that possessing the best AUC value does not necessarily translate to the best imbalanced performance. For instance, whilst the AUC for the HDTree is higher on the Letter, Pulsar and Landsat data sets, the G-Mean and the recall rate is lower. Whilst the HDTree has a better AUC value, the other algorithms, particularly C4.4, must be operating at points on the ROC curve which provide better results.
\begin{textblock}{170}(30,280)
\noindent Preprint submitted to 22nd International Conference on Pattern Recognition. Received December 19, 2013.
\end{textblock}
%\begin{table}
%\center
%\tabcolsep=0.11cm
%    \begin{tabular}{|c|c|c|c|c|c|}
%    \hline
%    \multicolumn{2}{|c}{Experiment} & \multicolumn{2}{|c}{Original Result} & \multicolumn{2}{|c|}{Our Result} \\\hline
%    Dataset & Algorithm & AUC & $\pm$ & AUC & Agreement \\\hline\hline
%    \multirow{3}{*}{Letter} & C4.5     & 0.990 & 0.004 & 0.933 & \xmark \\
% 										& C4.4     & 0.990 & 0.004 & 0.991 & \cmark \\
% 										& HDTree & 0.990 & 0.004 & 0.967 & \xmark \\\hline
%     \multirow{3}{*}{Pen} & C4.5     & 0.985 & 0.005 & 0.973 & \xmark \\
% 										& C4.4     & 0.985 & 0.005 & 0.986 & \cmark \\
% 										& HDTree & 0.992 & 0.002 & 0.987 & \xmark \\\hline
% 	 \multirow{3}{*}{Statlog (Landsat Satellite)} & C4.5 & 0.906 & 0.009 & 0.731 & \xmark \\
% 										& C4.4     & 0.906 & 0.009 & 0.912 & \cmark \\
% 										& HDTree & 0.911 & 0.007 & 0.915 & \cmark \\\hline								
%     \multirow{3}{*}{Statlog (Image Segmentation)} & C4.5 & 0.982 & 0.006 & 0.981 & \cmark \\
% 										& C4.4     & 0.982 & 0.006 & 0.987 & \cmark \\
%										& HDTree & 0.984 & 0.007 & 0.992 & \cmark \\\hline
%    \end{tabular}
%    \caption[Results]{Results reproduced from \cite{Cieslak:2008:cn}.}
%\end{table}

\begin{table}
\center
\tabcolsep=0.11cm
\scalebox{1.0}{
    \begin{tabular}{|c|c|c|c|c|c|}
    \hline
    Dataset & Algorithm & AUC & G-Mean & F-Score & Recall \\\hline\hline
    														     % AUC                  % G-MEAN               % F-Score          % RECALL
    \multirow{3}{*}{Pulsar} & C4.4     & .976 $\pm$ .03 &	 \textbf{.796 $\pm$ .03} & .628 $\pm$ .09 & \textbf{.634$\pm$ .14}\\
        														 % AUC                  % G-MEAN               % F-Score          % RECALL
 										& C4.5     & .867 $\pm$ .09 &	 .790 $\pm$ .07 & \textbf{.635 $\pm$ .06} & .624 $\pm$ .11\\
    														     % AUC                  % G-MEAN               % F-Score          % RECALL
 										& HDTree & \textbf{.992 $\pm$ .01} &	 .734 $\pm$ .04 & .555 $\pm$ .05 & .540 $\pm$ .06\\\hline
 										
    														     % AUC                  % G-MEAN               % F-Score          % RECALL
 	 \multirow{3}{*}{Skin}   & C4.4     & .999 $\pm$ .01 &	 .999 $\pm$ .01 & .998 $\pm$ .01 & .999 $\pm$ .01\\
    														     % AUC                  % G-MEAN               % F-Score          % RECALL
 										& C4.5     & .999 $\pm$ .01 &	 .999 $\pm$ .01 & .998 $\pm$ .01 & .999 $\pm$ .01\\
    														     % AUC                  % G-MEAN               % F-Score          % RECALL
 										& HDTree & .999 $\pm$ .01 &	 .999 $\pm$ .01 & .998 $\pm$ .01 & .998 $\pm$ .01\\\hline
 										
    														     % AUC                  % G-MEAN               % F-Score          % RECALL
 	 \multirow{3}{*}{MiniBoone} & C4.4 & .999 $\pm$ .01 &	 .999 $\pm$ .01 & .999 $\pm$ .01 & .999 $\pm$ .01\\
    														     % AUC                  % G-MEAN               % F-Score          % RECALL
 										& C4.5     & .999 $\pm$ .01 &	 .999 $\pm$ .01 & .999 $\pm$ .01 & .999 $\pm$ .01\\
    														     % AUC                  % G-MEAN               % F-Score          % RECALL
 										& HDTree & .999 $\pm$ .01 &	 .999 $\pm$ .01 & .999 $\pm$ .01 & .999 $\pm$ .01\\\hline	
 										
    														     % AUC                  % G-MEAN               % F-Score          % RECALL
    \multirow{3}{*}{Letter} & C4.4     & .965 $\pm$ .01 &	 \textbf{.913 $\pm$ .03} & .862 $\pm$ .03 & \textbf{.861 $\pm$ .05}\\
    														     % AUC                  % G-MEAN               % F-Score          % RECALL
 										& C4.5     & .933 $\pm$ .02 &	 .906 $\pm$ .04 & \textbf{.864 $\pm$ .05} & .844 $\pm$ .07\\
    														     % AUC                  % G-MEAN               % F-Score          % RECALL
 										& HDTree & \textbf{.967 $\pm$ .01} & .907 $\pm$ .02 & .860 $\pm$ .03 & .848 $\pm$ .05\\\hline
 										
    														     % AUC                  % G-MEAN               % F-Score          % RECALL
     \multirow{3}{*}{Magic} & C4.4 & \textbf{.888 $\pm$ .02} &	 .806 $\pm$ .02 & .883 $\pm$ .01 & .913 $\pm$ .03\\
    														     % AUC                  % G-MEAN               % F-Score          % RECALL
 										 & C4.5  & .856 $\pm$ .03 &	 \textbf{.809 $\pm$ .02} & \textbf{.884 $\pm$ .01} & \textbf{.914 $\pm$ .03}	\\
    														     % AUC                  % G-MEAN               % F-Score          % RECALL
 										 & HDTree & .884 $\pm$ .01 & .793 $\pm$ .02 & .853 $\pm$ .01 & .848 $\pm$ .02\\\hline
 										 
    														     % AUC                  % G-MEAN               % F-Score          % RECALL
     \multirow{3}{*}{Pen} & C4.4     & .986 $\pm$ .01 & \textbf{.965 $\pm$ .02} & .945 $\pm$ .02 & \textbf{.936 $\pm$ .05}\\
    														     % AUC                  % G-MEAN               % F-Score          % RECALL
 										& C4.5     & .968 $\pm$ .02 &	 .962 $\pm$ .02 & .945 $\pm$ .03 & .929 $\pm$ .03 \\
    														     % AUC                  % G-MEAN               % F-Score          % RECALL
 										& HDTree & \textbf{.988 $\pm$ .01} & \textbf{.965 $\pm$ .02} & \textbf{.948 $\pm$ .03} & \textbf{.936 $\pm$ .05}\\\hline
 										
    														     % AUC                  % G-MEAN               % F-Score          % RECALL
 	 \multirow{3}{*}{Statlog Landsat} & C4.4 & .912 $\pm$ .03 & \textbf{.716 $\pm$ .04} & .537 $\pm$ .07 & \textbf{.540 $\pm$ .07}\\
    														     % AUC                  % G-MEAN               % F-Score          % RECALL
 										& C4.5     & .731 $\pm$ .14 &	 .710 $\pm$ .07 & \textbf{.542 $\pm$ .10} & .528 $\pm$ .10\\
    														     % AUC                  % G-MEAN               % F-Score          % RECALL
 										& HDTree & \textbf{.916 $\pm$ .02} &	 .708 $\pm$ .04 & .541 $\pm$ .07 & .525 $\pm$ .05\\\hline	
 										
    														     % AUC                  % G-MEAN               % F-Score          % RECALL							
     \multirow{3}{*}{Statlog Image} & C4.4 & .987 $\pm$ .03 &	 .982 $\pm$ .02 & .971 $\pm$ .03 & .969 $\pm$ .05\\
    														     % AUC                  % G-MEAN               % F-Score          % RECALL
 										& C4.5     & .981 $\pm$ .02 &	 .980 $\pm$ .02 & .970 $\pm$ .03 & .964 $\pm$ .03\\
    														     % AUC                  % G-MEAN               % F-Score          % RECALL
 										& HDTree & \textbf{.992 $\pm$ .01} &	 \textbf{.987 $\pm$ .02} & \textbf{.978 $\pm$ .03} & \textbf{.978 $\pm$ .04}\\\hline
    \end{tabular}
    }
    \caption[Results]{Results obtained on static datasets using $5\times2$ CV.}
    \label{tab:static}
    \vspace{-2.0em}
\end{table} 
\begin{table*}[htdp]
\center
\footnotesize
\tabcolsep=0.1cm
\scalebox{1.0}{
    \begin{tabular}{|c|l|l|l|l|l|c|}
    \hline
    \multirow{2}{*}{Dataset} & Balance   & \multicolumn{4}{c|}{+1 : -10} & \multirow{2}{*}{Rank} \\ \cline{2-6}
    & Labelling (\%) & 10 & 50 & 75 & 100 &  \\ \hline\hline
    																	% Balance  +1 : -10
     \multirow{3}{*}{Pulsar}  & HTree &\cellcolor{RobGreen} .858/.926 &\cellcolor{RobGreen} .860/.904 &\cellcolor{RobGreen} .858/.901 &\cellcolor{RobGreen} .860/.905 & 2.5 \\
     													 				
     													 				% Balance  +1 : -10
     										& HD-VFDT &\cellcolor{RobGreen} .860/.922 &\cellcolor{RobGreen} .855/.923 &\cellcolor{RobGreen} .855/.933 &\cellcolor{RobGreen} .850/.920 & 1.25\\
     													 				
     													 				% Balance  +1 : -10
     										& GH-VFDT &\cellcolor{RobGreen} .858/.918 &\cellcolor{RobGreen} .852/.915 &\cellcolor{RobGreen} .851/.909 &\cellcolor{RobGreen} .850/.917 & 2.25\\\hhline{|-|-|-|-|-|-|-|}	%\cline{1-8}			
    													
    																	% Balance  +1 : -10
    \multirow{3}{*}{Skin}  & HTree & \cellcolor{RobGreen}.843/.911 & \cellcolor{RobGreen}.894/.958 & \cellcolor{RobGreen}.920/.969 & \cellcolor{RobGreen}.930/.973 & 1 \\
     																	% Balance  +1 : -10
     								  & HD-VFDT & \cellcolor{RobGreen}.737/.893 & \cellcolor{RobGreen}.746/.898 & \cellcolor{RobRed}.726/.877 & \cellcolor{RobYellow}.727/.888 & 2.75 \\
     																	% Balance  +1 : -10
    								& GH-VFDT  & \cellcolor{RobGreen}.815/.911 & \cellcolor{RobYellow}.824/.904 & \cellcolor{RobYellow}.835/.913 & \cellcolor{RobYellow}.816/.914 & 1.75  \\\hhline{|-|-|-|-|-|-|-|}	%\cline{1-8}		
    
   																		 % Balance  +1 : -10
     \multirow{3}{*}{MiniBoone} & HTree  & \cellcolor{RobGreen}.954/.958 & \cellcolor{RobGreen}.993/.993 & \cellcolor{RobGreen}.992/.993 & \cellcolor{RobGreen}.992/.992 & 1 \\  
    																	% Balance  +1 : -10
    											& HD-VFDT& \cellcolor{RobGreen}.889/.913 &\cellcolor{RobGreen} .859/.786 & \cellcolor{RobGreen}.882/.912 & \cellcolor{RobGreen}.859/.818 & 2 \\
    																	% Balance  +1 : -10
    											& GH-VFDT & \cellcolor{RobYellow}.568/.398 & \cellcolor{RobGreen}.666/.450 & \cellcolor{RobYellow}.661/.497 & \cellcolor{RobGreen}.760/.665 & 3 \\
    \hline
    \end{tabular}
    \begin{tabular}{|l|l|l|l|l|c|}
    \hline
    Balance  & \multicolumn{4}{c|}{+1 : -100} & \multirow{2}{*}{Rank} \\ \cline{1-5}
    Labelling (\%) & 10 & 50 & 75 & 100 & \\ \hline\hline
    																	% Balance  +1 : -100
      HTree &\cellcolor{RobGreen} .673/.876 &\cellcolor{RobYellow} .752/.854 &\cellcolor{RobYellow} .753/.852 &\cellcolor{RobYellow} .767/.859 & 2.75\\     													 				
     													 				% Balance  +1 : -100
     HD-VFDT &\cellcolor{RobGreen} .457/.835 &\cellcolor{RobGreen} .472/.929 &\cellcolor{RobGreen} .529/.919 &\cellcolor{RobGreen} .493/.931 & 1.5\\		 				
     													 				% Balance  +1 : -100
    GH-VFDT  &\cellcolor{RobGreen} .518/.903 &\cellcolor{RobGreen} .483/.920 &\cellcolor{RobGreen} .536/.916 &\cellcolor{RobGreen} .552/.916 & 1.75 \\\hhline{|-|-|-|-|-|-|}	%\cline{1-8}				
    													
    																	% Balance  +1 : -100
    HTree  & \cellcolor{RobYellow}.472/.671 & \cellcolor{RobYellow}.486/.615 & \cellcolor{RobYellow}.541/.655 & \cellcolor{RobYellow}.574/.693 & 3\\
     																	% Balance  +1 : -100
    HD-VFDT & \cellcolor{RobGreen}.277/.887 & \cellcolor{RobGreen}.292/.885 & \cellcolor{RobGreen}.319/.903 & \cellcolor{RobGreen}.263/.892 & 2\\
     																	% Balance  +1 : -100
   GH-VFDT  &\cellcolor{RobGreen}.399/.898 & \cellcolor{RobGreen}.478/.915 & \cellcolor{RobGreen}.441/.904 & \cellcolor{RobGreen}.430/.900 & 1 \\\hhline{|-|-|-|-|-|-|}	%\cline{1-8}	
    
   																		 % Balance  +1 : -100
    HTree  & \cellcolor{RobGreen}.918/.932 &\cellcolor{RobGreen} .964/.972 &\cellcolor{RobGreen} .972/.978 &\cellcolor{RobGreen} .980/.982 & 1\\  
    																	% Balance  +1 : -100
    HD-VFDT & \cellcolor{RobGreen}.874/.866 &\cellcolor{RobGreen} .791/.857 &\cellcolor{RobGreen} .758/.854 &\cellcolor{RobGreen} .876/.949 & 2\\
    																	% Balance  +1 : -100
    GH-VFDT  & \cellcolor{RobGreen}.742/.597&\cellcolor{RobYellow} .547/.395 &\cellcolor{RobGreen} .623/.498 &\cellcolor{RobGreen} .511/.459 & 3\\
    \hline
    \end{tabular}
    }
    \caption[Results]{F-score/G-Mean results for the \textit{HoeffdingTree} (HTree), \textit{Hellinger Distance Tree} (HD-VFDT) \cite{Cieslak:2008:cn} and \textit{GH-VFDT}  classifiers, trained before classifying a stream. Imbalances ranging from +1:-10 to +1:-100, significance tests used $\alpha=0.01$.}
    \vspace{-2em}
\label{tab:resultsOne}
\end{table*}

\begin{table*}[htdp]
\center
\footnotesize
\tabcolsep=0.1cm
\scalebox{1.0}{
    \begin{tabular}{|c|l|l|l|l|l|c|}
    \hline
    \multirow{2}{*}{Dataset} & Balance   & \multicolumn{4}{c|}{+1 : -1,000} & \multirow{2}{*}{Rank}  \\ \cline{2-6}
    & Labelling (\%) & 10 & 50 & 75 & 100 &  \\ \hline\hline
    																	% Balance  +1 : -1,000
     \multirow{3}{*}{Pulsar}  & HTree &\cellcolor{RobYellow} .372/.746 &\cellcolor{RobYellow} .576/.785 &\cellcolor{RobYellow} .581/.795 &\cellcolor{RobYellow} .610/.781 & 3 \\
     													 				
     													 				% Balance  +1 : -1,000
     										& HD-VFDT &\cellcolor{RobGreen} .106/.924 &\cellcolor{RobGreen} .107/.929 &\cellcolor{RobGreen} .106/.924 &\cellcolor{RobGreen} .100/.934 & 1 \\
     													 				
     													 				% Balance  +1 : -1,000
     										& GH-VFDT &\cellcolor{RobGreen} .137/.910 &\cellcolor{RobGreen} .131/.915 &\cellcolor{RobGreen} .128/.905 &\cellcolor{RobGreen} .125/.928 & 2 \\\hhline{|-|-|-|-|-|-|-|}	%\cline{1-8}					
    													
    																	% Balance  +1 : -1,000
    \multirow{3}{*}{Skin}  & HTree & \cellcolor{RobYellow}.137/.532 &\cellcolor{RobYellow} .106/.302 &\cellcolor{RobYellow}  .069/.225 &\cellcolor{RobYellow}  .087/.239 & 3 \\
     																	% Balance  +1 : -1,000
     								  & HD-VFDT & \cellcolor{RobGreen}.043/.802 &\cellcolor{RobGreen} .047/.891 &\cellcolor{RobGreen} .051/.893 &\cellcolor{RobGreen} .046/.882 & 2 \\
     																	% Balance  +1 : -1,000
    								& GH-VFDT  & \cellcolor{RobGreen}.082	/.911 &\cellcolor{RobGreen} .082/.921 &\cellcolor{RobGreen} .080/.922 &\cellcolor{RobGreen} .080/.908 & 1 \\\hhline{|-|-|-|-|-|-|-|}	%\cline{1-8}	
    
   																		 % Balance  +1 : -1,000
     \multirow{3}{*}{MiniBoone} & HTree  &\cellcolor{RobGreen} .576	/.617 &\cellcolor{RobGreen} .918/.951 &\cellcolor{RobGreen} .812/.862 &\cellcolor{RobGreen} .907/.943 & 1.5 \\  
    																	% Balance  +1 : -1,000
    											& HD-VFDT &\cellcolor{RobGreen} .284/.842 &\cellcolor{RobGreen} .650/.837 &\cellcolor{RobGreen} .206/.776 &\cellcolor{RobGreen} .455/.944 & 1.5 \\
    																	% Balance  +1 : -1,000
    											& GH-VFDT &\cellcolor{RobGreen} .293/.469 &\cellcolor{RobGreen} .709/.656 &\cellcolor{RobYellow} .332/.200 &\cellcolor{RobYellow} .331/.300  & 3 \\
    \hline
    \end{tabular}
   \begin{tabular}{|l|l|l|l|l|c|c|}
    \hline
    Balance  & \multicolumn{4}{c|}{+1 : -10,000} & \multirow{2}{*}{Rank}  \\ \cline{1-5}
    Labelling (\%) & 10 & 50 & 75 & 100 &   \\ \hline\hline
    																	% Balance  +1 : -10,000
    HTree &\cellcolor{RobYellow} .194/	.544 &\cellcolor{RobYellow} .114/.289 &\cellcolor{RobYellow} .159/.334 &\cellcolor{RobYellow} .170/.331 & 3 \\
     													 				
     													 				% Balance  +1 : -10,000
    HD-VFDT &\cellcolor{RobGreen} .014/.921 &\cellcolor{RobGreen} .013/.928 &\cellcolor{RobGreen} .011/.938 &\cellcolor{RobGreen} .016/.919 & 1 \\
     													 				
     													 				% Balance  +1 : -10,000
    GH-VFDT &\cellcolor{RobGreen} .028	/.904 &\cellcolor{RobGreen} .016/.920 &\cellcolor{RobGreen} .020/.917 &\cellcolor{RobGreen} .026/.886 & 2 \\\hhline{|-|-|-|-|-|-|}	%\cline{1-8}				
    													
    																	% Balance  +1 : -10,000
    HTree &\cellcolor{RobYellow} .018	/.511 &\cellcolor{RobYellow} .017/.219 &\cellcolor{RobYellow} .011/.095 &\cellcolor{RobYellow} .021/.141 & 3 \\
     																	% Balance  +1 : -10,000
    HD-VFDT &\cellcolor{RobGreen} .004	/.871 &\cellcolor{RobGreen} .005/.914 &\cellcolor{RobGreen} .005/.908 &\cellcolor{RobGreen} .006/.869 & 1.75 \\
     																	% Balance  +1 : -10,000
    GH-VFDT  &\cellcolor{RobGreen} .008/.895 &\cellcolor{RobGreen} .009/.892 &\cellcolor{RobGreen} .009/.917 &\cellcolor{RobGreen} .009/.916 & 1.25 \\\hhline{|-|-|-|-|-|-|}	%\cline{1-8}	
    
   																		 % Balance  +1 : -10,000
     HTree  &\cellcolor{RobGreen} .604/.672 &\cellcolor{RobGreen} .586/.667 &\cellcolor{RobGreen} .651/.830 &\cellcolor{RobGreen} .649/.707 & 2 \\  
    																	% Balance  +1 : -10,000
    HD-VFDT &\cellcolor{RobGreen} .021/.844 &\cellcolor{RobGreen} .039/.792 &\cellcolor{RobGreen} .048/.875 &\cellcolor{RobGreen} .328/.853 & 1 \\
    																	% Balance  +1 : -10,000
    GH-VFDT &\cellcolor{RobGreen} .667/.500 &\cellcolor{RobGreen} .260/.447 &\cellcolor{RobGreen} .603/.494 &\cellcolor{RobGreen} .302/.489 & 3 \\
    \hline
    \end{tabular}
    }
    \caption[Results]{F-score/G-Mean results for the \textit{HoeffdingTree} (HTree), \textit{Hellinger Distance Tree} (HD-VFDT) \cite{Cieslak:2008:cn} and \textit{GH-VFDT}  classifiers, trained before classifying a stream. Imbalances ranging from +1:-1,000 to +1:-10,000, significance tests used $\alpha=0.01$.}
    \vspace{-1.5em}
\label{tab:resultsTwo}
\end{table*}
In the tables that follow the results for the streamed test scenarios are given. In these tables the significance of each result is indicated using colour coding. Different colours indicate statistically different results at the level $\alpha=0.01$. Here green indicates the best performing and red the worst i.e. green~$>$~yellow~$>$~red. Table \ref{tab:resultsOne} shows the results obtained on data streams with imbalances ranging from +1:-10 to +1:-100 with varied levels of labelling. For the least imbalanced streams (imbalance +1:-10), we see that for pulsar data there was no significant difference between the algorithms at the $\alpha=0.01$ level. However there were significant differences observed during testing on the Skin and MiniBoone datasets. Here the VFDT consistently outperforms the two Hellinger based approaches. The performance of the GH-VFDT in particular worsens on Skin data as the labelling increases. The HD-VFDT is also affected since its performance drops when labelling reaches 75\%. Crucially however, as the imbalance worsens to +1:-100 this situation reverses. The two Hellinger based approaches maintain their high G-Mean values, whilst for the VFDT this metric drops. The Hellinger algorithms now perform significantly better than the VFDT on both the Pulsar and Skin datasets at $\alpha=0.01$. This trend continues as the imbalance worsens further as shown in Table \ref{tab:resultsTwo}. Here we see that the G-Mean of the VFDT has effectively dropped from $.905$ for the least imbalanced pulsar stream with 100\% labelling, to $.331$ for the most imbalanced with the same labelling. Both Hellinger approaches by contrast have maintained consistent G-Mean values regardless of the imbalance. This is reflected in the minority class recall rates for the three classifiers. The HD-VFDT and GH-VFDT return more than double the true positives of the VFDT on pulsar data (see Table \ref{tab:Recall}).  The MiniBoone results are an exception to this trend, since all three algorithms perform similarly here throughout.
\begin{table}[htdp]
\center
\footnotesize
\tabcolsep=0.1cm
    \begin{tabular}{|l|l|l|l|l|c|}
    \hline
   Balance   & \multicolumn{4}{c|}{+1 : -10,000} \\ \cline{1-5}
   Labelling (\%) & 10 & 50 & 75 & 100  \\ \hline\hline
   HTREE 		& .307/.001 & .087/.001 & .116/.001 & .121/.001 \\    										 				
   HD-VFDT 	& \textbf{.863/.017} & \textbf{.876/.016} & \textbf{.895/.017} & \textbf{.860/.016}\\
   GH-VFDT 	& .829/.014 & .861/.017 & .855/.013 & .801/.012\\
    \hline
    \end{tabular}
    \caption[]{Recall/false positive rate on pulsar data with an imbalance of +1:-10,000 and 10\% labelled data.}
    \label{tab:Recall}
    \vspace{-2em}
\end{table}

In summary these results indicate that for streams possessing class imbalances greater than +1:-10, the use of algorithms such as the GH-VFDT and HD-VFDT can result in statistically significant increases in the minority class recall rate over standard stream classifiers such as the VFDT. Of the Hellinger distance based approaches described in this paper, our stream implementation of the HDTree algorithm from \cite{Cieslak:2008:cn} was the best performer. It typically obtained the highest G-Mean values, and significantly improved recall rates over the VFDT. Our algorithm GH-VFDT, also significantly improved recall rates over the VFDT. Whilst it did not achieve recall rates quite as high as the HD-VFDT, its false positive rate was in fact lower. Crucially the difference between these algorithms on the most imbalanced streams was not significant at $\alpha=0.01$ (on all but the MiniBoone data). The GH-VFDT and HD-VFDT also achieved G-Mean values and recall rates which were higher than those observed for each of the static classifiers described in Section\ref{sec:Experiments_A}. Thus we argue that the GH-VFDT is a viable classifier for imbalanced streams particularly when the memory required by the HD-VFDT hinders its use.

\section{Related Work}
\label{sec:Related}
\begin{textblock}{170}(30,280)
\noindent Preprint submitted to 22nd International Conference on Pattern Recognition. Received December 19, 2013.
\end{textblock}
The problems associated with learning from an imbalanced class distribution, are typically tackled using one of three approaches. \textit{Data level} approaches seek to modify data sets in order to rebalance the class distribution directly. \textit{Algorithm level} approaches focus on modifying algorithms rather than datasets, to achieve improved performance on the minority class examples. \textit{Cost sensitive} approaches on the other hand utilise different cost matrices describing the costs associated with misclassifying a particular data instance, reflecting the fact that it is usually more costly to misclassify rare class instances. There are also many hybrids of these approaches, particularly of those that mix data and algorithm level approaches in some way. However practical efforts at solving the imbalanced learning problem have focused on developing new data level sampling techniques. Some representative approaches include One-sided selection (OSS) \cite{Kubat97addressingthe}, Wilson’s Editing (WE) \cite{Wilson:1972:dl,Barandela:2004:vr}, the Synthetic Minority Over-sampling Technique (SMOTE) \cite{Chawla:2002:bk}, Borderline SMOTE \cite{Han:2005:wy}, and ADASYN \cite{Haibo:2008:yb}.%Akbani et al. make the case that such undersampling can often omit useful information which can lead to sub-optimal decision boundaries for SVMs \cite{Akbani:2004:sk}.

More recently some have begun investigating how these perform on streaming data. Nguyen et al. \cite{Nguyen:2012:ce} for instance compared over-sampling and under-sampling techniques in the context of streamed data. Their results suggest that whilst under-sampling performs better than over-sampling at smaller training set sizes, performance converges as the training set becomes larger (i.e. as more streamed data is used to train a classifier). Based upon these results the authors propose a new approach, multiple random under-sampling (MRUS), which generates $m$ random undersamplings of the stream, each of which is used to train a separate classifier forming an ensemble. In \cite{Haibo:2011:cs} Chen et al. developed their recursive weighted ensemble approach (REA) for classifying non-stationary imbalanced data streams. REA adaptively pushes minority class examples into the current data chunk to explicitly balance the class distribution. The Stream ensemble framework developed in \cite{Gao:2008:bd} attempts to mitigate similar problems by combining an ensemble classifier with sampling approaches. Each classifier in the ensemble is trained on a data set containing all the positive instances seen in the stream, and a unique subset of negative instances undersampled from the stream. The approach is designed to operate under a batch model. 

\begin{textblock}{170}(30,280)
\noindent Preprint submitted to 22nd International Conference on Pattern Recognition. Received December 19, 2013.
\end{textblock}
\section{Conclusion}
In this paper we have presented a new classification algorithm for imbalanced data streams called GH-VFDT. Through an empirical investigation we have demonstrated that the algorithm can effectively improve minority class recall rates on imbalanced data, with similar levels of performance to the algorithm described in \cite{Cieslak:2008:cn}, whilst returning fewer false positives. A natural extension to this work would be to modify the decision tree algorithm so that rather than predicting the majority class at a tree leaf, we predict the class that possesses the maximal summed affinity to the features possessed by an instance. Furthermore, we would like to change how split points are calculated at nodes, such that the best split is obtained by choosing the point with the lowest Hellinger distance variance.
%In this paper we have presented a new classification algorithm for imbalanced data streams called GH-VFDT. Through an empirical investigation we have demonstrated that the algorithm can effectively improve minority class recall rates on imbalanced data, whilst also achieving similar levels of performance to the original algorithm described in \cite{Cieslak:2008:cn}, and better performance than the VFDT upon which it is based \cite{Hulten:2001:MTD}, whilst returning fewer false positives. A natural extension to this work would be to modify the decision tree algorithm so that rather than predicting the majority class at a tree leaf, we predict the class that possesses the maximal summed affinity to the features possessed by an instance. Furthermore, we would like to change how split points are calculated at nodes, such that the best split is obtained by choosing the point with the lowest Hellinger distance variance.
\section*{Acknowledgment}
This work was supported by grant EP/I028099/1 for the University of Manchester Centre for Doctoral Training in Computer Science, from the UK Engineering and Physical Sciences Research Council (EPSRC). Experiments were carried out upon data obtained by the Parkes Observatory, funded by the Commonwealth of Australia and managed by the CSIRO. %Commonwealth Scientific and Industrial Research Organisation (CSIRO).



%\bibliographystyle{IEEEtran}
%\begin{thebibliography}{1}


%\end{thebibliography}

\bibliographystyle{ieeetr.bst}

\end{document}